\documentclass[letterpaper, 10 pt, conference]{ieeeconf}  

\IEEEoverridecommandlockouts                              

\usepackage{graphics} 
\usepackage{epsfig} 
\usepackage{times} 
\usepackage{amsmath} 
\usepackage{amssymb}  
\usepackage{adjustbox}
\usepackage{multirow}
\usepackage{hhline}
\usepackage{array}
\usepackage{makecell}
\usepackage{bibentry}
\usepackage{epstopdf,cite,subfigure}
\usepackage{xcolor}
\usepackage{color}
\usepackage{bbding}
\usepackage{pifont}

\title{\LARGE \bf
D-Align: Dual Query Co-attention Network for 3D  Object Detection Based on Multi-frame Point Cloud Sequence
}

\author{Junhyung Lee$^{1}$, Junho Koh$^{2}$, Youngwoo Lee$^{2}$ and Jun Won Choi$^{2}$
\thanks{$^{1}$Department of Future Mobility, Hanyang University, Seoul, 04763, Korea}
\thanks{$^{2}$Department of Electrical Engineering, Hanyang University, Seoul, 04763, Korea}
\thanks{{\tt\footnotesize \{junhyunglee,jhkoh,youngwoolee\}@spa.hanyang.ac.kr}}
\thanks{{\tt\footnotesize junwchoi@hanyang.ac.kr}}
}

\begin{document}

\maketitle
\thispagestyle{empty}
\pagestyle{empty}

\begin{abstract}
LiDAR sensors are widely used for 3D object detection in various mobile robotics applications. LiDAR sensors continuously generate point cloud data in real-time. Conventional 3D object detectors detect objects using a set of points acquired over a fixed duration. However, recent studies have shown that the performance of object detection can be further enhanced by utilizing spatio-temporal information obtained from point cloud sequences. In this paper, we propose a new 3D object detector, named {\it D-Align}, which can effectively produce strong bird's-eye-view (BEV) features by aligning and aggregating the features obtained from a sequence of point sets. The proposed method includes a novel dual-query co-attention network that uses two types of queries, including target query set (T-QS) and support query set (S-QS), to update the features of target and support frames, respectively. D-Align aligns S-QS to T-QS based on the temporal context features extracted from the adjacent feature maps and then aggregates S-QS with T-QS using a gated attention mechanism. The dual queries are updated through multiple attention layers to progressively enhance the target frame features used to produce the detection results. Our experiments on the nuScenes dataset show that the proposed D-Align method greatly improved the performance of a single frame-based baseline method and significantly outperformed the latest 3D object detectors.
\end{abstract}

\section{INTRODUCTION}
LiDAR is a widely used sensor modality for perception tasks in various mobile robotics applications. LiDAR sensors generate point cloud data corresponding to observations of laser reflections from the surfaces of objects. Point cloud data are particularly useful for 3D object detection tasks, which involve estimating objects' locations in 3D coordinate systems.
Recently, LiDAR-based 3D object detection has advanced rapidly with the adoption of deep neural networks to extract features from point cloud data. 

\begin{figure}[t]
    \centering
    \begin{subfigure}[]{\includegraphics[width=0.82\columnwidth]{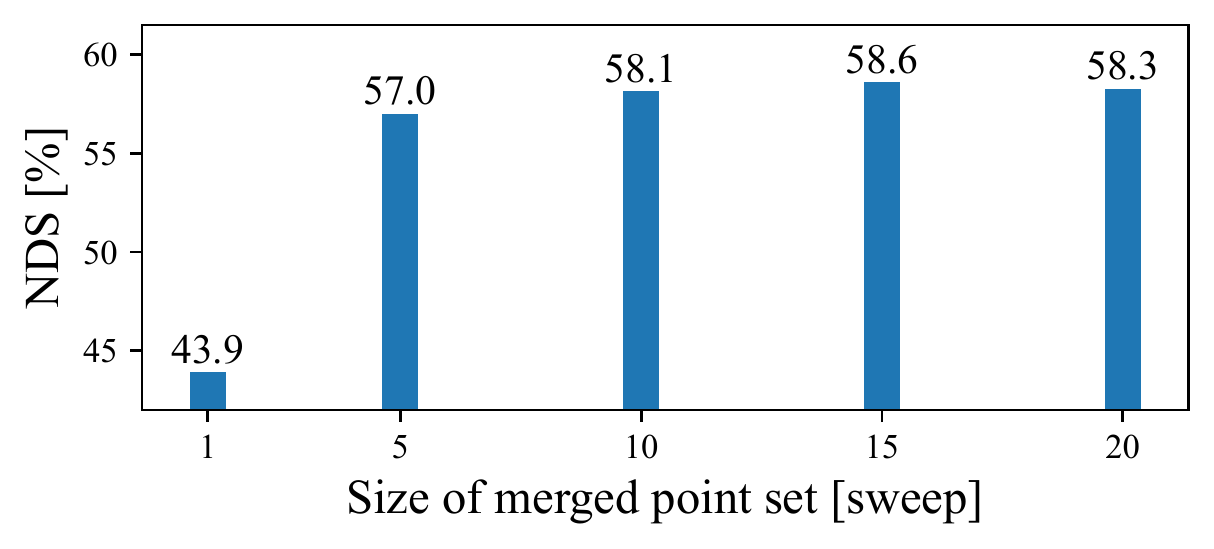}}
    \end{subfigure}
    \vspace{2.5mm}
    \begin{subfigure}[]{\includegraphics[width=0.82\columnwidth]{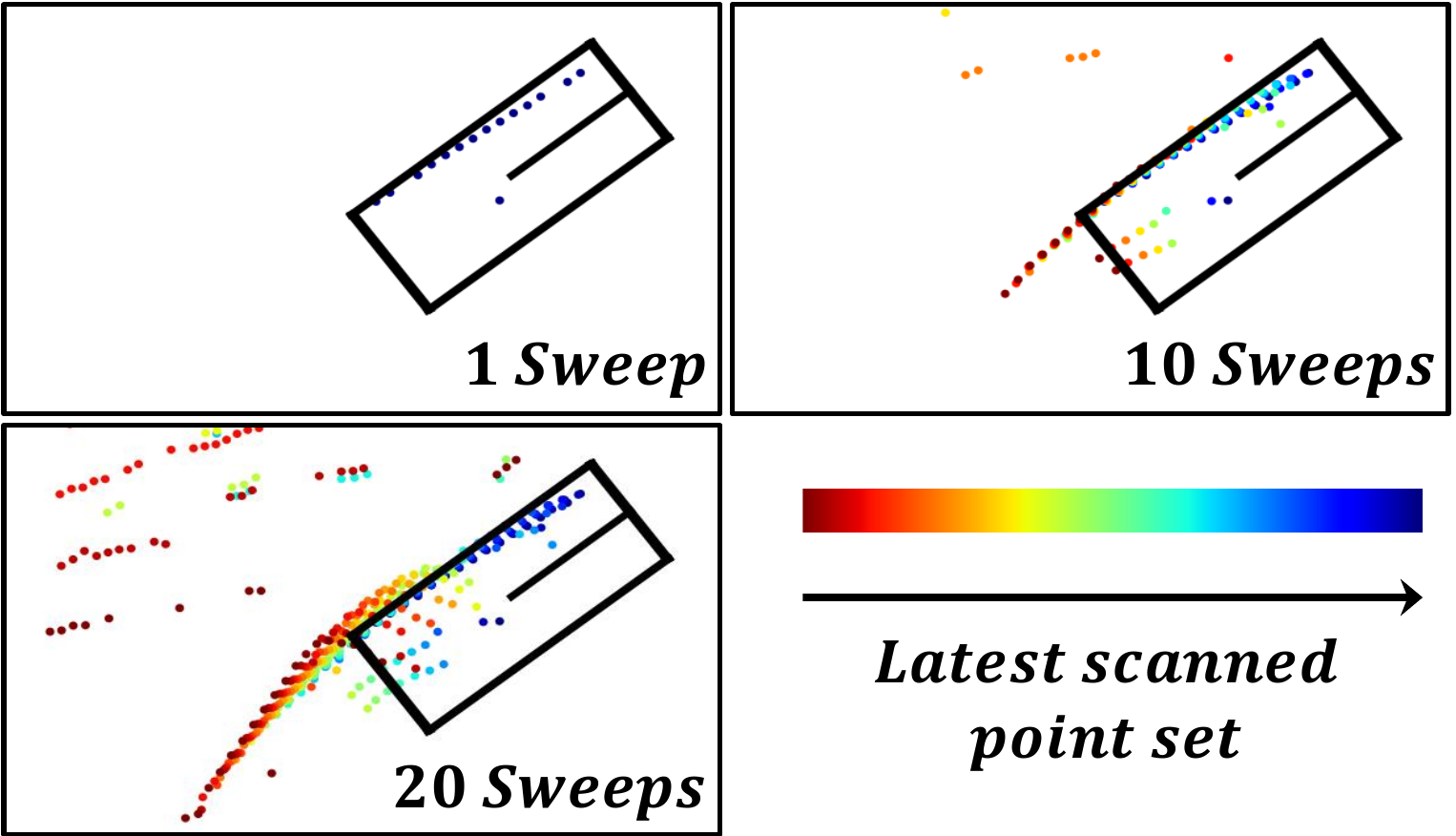}}
    \end{subfigure}
    \caption {\textbf{Effects of using larger LiDAR point sets for 3D object detection:} (a) Evaluation results of the PointPillars \cite{pointpillars} method on the nuScenes \cite{nuscenes} validation set. NDS denotes the nuScenes detection score. The NDS performance improvement of the 3D object detector quickly diminished as the size of LiDAR point set increased. (b) Object motion leads to dynamic changes in the distribution of point clouds  as more LiDAR points are merged.}
    \label{Motivation_NumSweeps}
\end{figure}

Conventional 3D object detectors operate on a set of LiDAR points, called {\it point clouds}, which are acquired by a fixed number of consecutive laser scans.   
The geometrical distribution of point clouds is used to detect objects in 3D space.  
In practical applications, LiDAR sensors continuously scan their surroundings to generate point cloud sequences in real time. Therefore, the quality of point cloud data can be improved by using more than one set of point clouds for 3D object detection.
One na\"ive approach is to merge multiple consecutive sets of point clouds acquired in the fixed duration and use the larger set of points as input to a 3D object detection model \cite{nuscenes}. This effectively improves the density of the point clouds and thus improves performance, as shown in Fig. \ref{Motivation_NumSweeps} (a). However, this performance improvement quickly diminishes with increasing numbers of point clouds merged because the distribution of the set of points changes across multiple scans. Fig. \ref{Motivation_NumSweeps} (b) shows that a moving object exhibits a dynamically changing point distribution even after compensating for the ego vehicle's motion that occurred in the duration.
Therefore, more sophisticated 3D object detection models that capture the dynamic temporal structure of LiDAR points are needed to improve accuracy.

Several recent studies have explored methods to utilize temporal information existing in long sequences of point clouds for 3D object detection. In this study, we refer to these detectors as  \textit{ 3D multi-frame object detectors} (3D-MOD). Various architectures have been proposed for 3D-MOD \cite{LSTM_LiDARVOD,stgnn, 3dman,3dvid,tctr}.
These methods extracted geometric features from each set of points (called a {\it frame}) within which the point distribution did not change much and modeled their temporal variations to improve performance.
In LSTM-TOD\cite{LSTM_LiDARVOD} and 3DVID \cite{3dvid}, the multiple bird's-eye-view (BEV) feature maps extracted from the point cloud sequence were combined with the variant of the recurrent module, i.e., ConvLSTM\cite{ConvLSTM} and ConvGRU\cite{ConvGRU}.
 TCTR \cite{tctr} exploited temporal-channel relations over multiple feature maps using an encoder-decoder structure \cite{Transformer}.
3D-MAN \cite{3dman} aggregated box-level features with a memory bank that containing sequential temporal view information.

In this paper, we present a new 3D-MOD method called {\it Dual-Query Align (D-Align)}, which can produce robust spatio-temporal BEV representations using a multi-frame point cloud sequence. We propose a novel {\it dual query co-attention network} that employs two types of queries: {\it target query set} (T-QS) and {\it support query set} (S-QS) to facilitate the co-attention to both the target and support frame features.  T-QS and S-QS serve to carry the target frame features and the support frame features, which are continuously enhanced through multiple layers of attention.

In each attention layer, the dual queries are updated in two steps. First, the {\it inter-frame deformable alignment} network (IDANet) aligns S-QS to T-QS using deformable attention. The deformable attention mechanism is applied to S-QS with mask offsets and weights determined by multi-scale temporal context features generated from two adjacent BEV feature maps. This step updates S-QS first. Next, the  {\it inter-frame gated aggregation} network (IGANet) aggregates S-QS and T-QS using a gated attention network \cite{GAM}. The aggregated query features finally update T-QS. After going through multiple attention layers, D-Align produces the improved BEV features of the target frame for 3D object detection.

We evaluate the proposed D-Align on the widely used
public nuScenes dataset \cite{nuscenes}. Our experimental results show that the proposed method improves the 3D detection baseline by significant margins and outperforms the latest LiDAR-based state-of-the-art (SOTA) 3D object detectors.

The contributions of our paper are summarized as follows.
\begin{itemize}
    \item 
    We propose an enhanced 3D object detection architecture, D-Align, which exploits temporal structure in point cloud sequences.
    We devise a novel dual query co-attention network that transforms the dual queries S-QS and T-QS through successive operations of feature alignment and feature aggregation. This co-attention mechanism allows attending to both the support and target frame features to gather useful spatio-temporal information from multiple frames of the point data.
    
    \item We design the temporal context-guided deformable attention to achieve inter-frame feature alignment. Our deformable attention mechanism differs from the original model proposed in \cite{def-detr} in that the attention mask is adjusted by the motion context obtained from two adjacent BEV feature maps. Our analysis shows that the use of such motion features contributed significantly to the overall detection performance.
    
    \item Our codes will be publicly released.
  
\end{itemize}

\section{RELATED WORK}
3D object detection techniques based on a single point set \cite{pointpillars,voxelnet,second,centerpoint,AFDet} have advanced rapidly since deep neural networks have been adopted to encode irregular and unordered point clouds. However, the performance of these 3D object detectors remains limited because they do not exploit temporal information in sequence data.

To date, several 3D-MOD methods have been proposed, which used point cloud sequences to perform 3D object detection \cite{LSTM_LiDARVOD,3dvid,tctr,3dman}. These methods explored ways to represent time-varying features obtained from long sequence of point clouds. In \cite{LSTM_LiDARVOD,3dvid,tctr}, LIDAR features obtained from multiple point cloud frames were combined to exploit temporal information in sequence data. LSTM-TOD \cite{LSTM_LiDARVOD} produced a spatio-temporal representation of point cloud sequences using a 3D sparse ConvLSTM\cite{ConvLSTM} modified to encode the point cloud sequence data. 3DVID \cite{3dvid} improved the conventional ConvGRU by adopting a transformer attention mechanism to exploit the spatio-temporal coherence of point cloud sequences. TCTR \cite{tctr} explored channel-wise temporal correlations among consecutive frames and decoded spatial information using Transformer. 3D-MAN \cite{3dman} stored 3D proposals and 3D spatial features obtained from a single-frame 3D detector in a memory bank. Then, to integrate the local features of objects extracted from each frame, the method explored spatio-temporal relationships among the proposals.

The proposed D-Align differs from these methods in that a novel dual query co-attention architecture is introduced to utilize spatio-temporal information obtained from point cloud sequences. The proposed method effectively aligns and aggregates multi-frame features by refining the dual query sets through multiple attention layers.

\begin{figure*}[t]
	\centering
        \centerline{\includegraphics[width=0.95\textwidth]{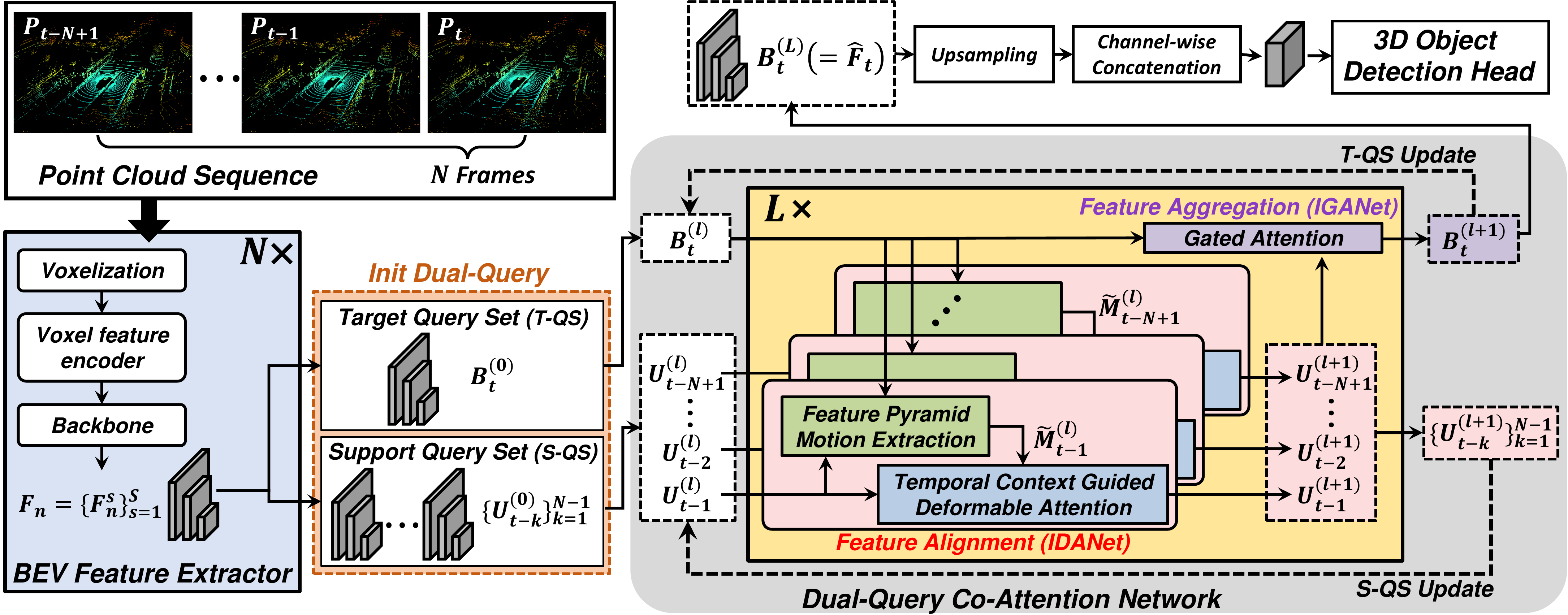}}
    	\caption {\textbf{Overall architecture of the proposed D-Align:} D-Align employs a dual query co-attention network to align and aggregate multi-frame BEV features. T-QS and S-QS are dual queries used to refine the features of the target and support frames, respectively. For each attention layer, S-QS is refined by applying the temporal context-guided deformable attention to S-QS for inter-frame feature alignment. T-QS is also updated by aggregating the updated S-QS with T-QS via the gated attention network. These updated dual queries are used as input queries for the next attention layer. }
	\label{overall}
\end{figure*}

\section{PROPOSED METHOD}
Fig. \ref{overall} shows the overall structure of D-Align. It consists of three main blocks, including 1) BEV feature extractor, 2) {\it dual-query co-attention network} (DUCANet), and 3) 3D object detection head.
The set of points acquired by the LiDAR sensor over the duration of $T$ seconds is called a frame. Ego-motion compensation \cite{nuscenes} is applied to the points within each frame. D-Align takes the sequence of point clouds $\{\textit{P}_{n}\}_{n=t-N+1}^{t}$ in $N$ successive frames as an input, where $\textit{P}_{n}$ denotes the point set obtained in the $n$th frame.  The frame $t$ is called a target frame because we aim to detect objects for the frame $t$. The remaining frames are called support frames.

\subsection{Overview}
The BEV feature extractor produces the BEV feature maps $\{\textbf{\textit{F}}_{t-k}\}_{k=0}^{N-1}$ by using a grid-based backbone network for each frame \cite{pointpillars}, \cite{second}. This backbone network produces the feature maps of $\it{S}$ scales, i.e., $\textbf{\textit{F}}_{n}=\{\textbf{\textit{F}}^s_{n}\}_{s=1}^{S}$. Next, DUCANet produces the enhanced target frame BEV features by applying the dual-query co-attention mechanism to the multi-frame features $\{\textbf{\textit{F}}_{t-k}\}_{k=0}^{N-1}$. DUCANet maintains two query sets, T-QS and S-QS. T-QS serves to store the target frame features transformed by the attention mechanism, whereas S-QS stores the support frame features. We denote T-QS and S-QS as $\textbf{\textit{B}}^{(l)}_{t}$ and $\{\textbf{\textit{U}}^{(l)}_{t-k}\}_{k=1}^{N-1}$, respectively, where $l$ is the index for the attention layers. 
In the first attention layer, T-QS $\textbf{\textit{B}}^{(0)}_{t}$ is initialized by $\textbf{\textit{F}}_{t}$. Similarly, S-QS $\{\textbf{\textit{U}}^{(0)}_{t-k}\}_{k=1}^{N-1}$ is initialized with $\{\textbf{\textit{F}}_{t-k}\}_{k=1}^{N-1}$. 
For each attention layer, T-QS and S-QS are refined by aligning and aggregating the multi-frame features through IDANet and IGANet.

First, IDANet aligns S-QS to T-QS by applying the temporal context-guided deformable attention to S-QS. Deformable attention \cite{def-detr} has been widely used to spatially transform input features adequately for a given task.
The {\it feature pyramid motion extraction network} (FPMNet) computes the temporal context features, $\tilde{\textbf{\textit{M}}}^{(l)}_{t-k}$ from two query features $\textbf{\textit{B}}^{(l)}_t$ and $\textbf{\textit{U}}^{(l)}_{t-k}$. Then, IDANet determines the deformable mask using the temporal context features $\{\tilde{\textbf{\textit{M}}}^{(l)}_{t-k}\}_{k=1}^{N-1}$. Using the outputs of the deformable attention,  S-QS is updated with $\{\textbf{\textit{U}}^{(l+1)}_{t-k}\}_{k=1}^{N-1}$.
Next, IGANet aggregates S-QS refined by IDANet with T-QS   using the gated attention network proposed in \cite{GAM}.
This step updates T-QS with $\textbf{\textit{B}}^{(l+1)}_{t}$. Finally, these refined dual query sets $\{\textbf{\textit{U}}^{(l+1)}_{t-k}\}_{k=1}^{N-1}$ and $\textbf{\textit{B}}^{(l+1)}_{t}$ are used as input queries for the next attention layer.

After going through $L$ attention layers, DUCANet comes up with the final T-QS, $\textbf{\textit{B}}^{(L)}_{t}$. This T-QS is used as the multi-scale BEV feature maps $\hat{\textbf{\textit{F}}}_{t} \left(= \{ \hat{\textbf{\textit{F}}}^{s}_{t} \}_{s=1}^{S} \right)$ to detect objects at frame $t$. After re-sizing these multi-scale features to have the same scale and concatenating them channel-wise, a 3D object detection head is applied to produce the final 3D detection results.

\begin{figure}[t]
	\centering
        \centerline{\includegraphics[width=0.44\textwidth]{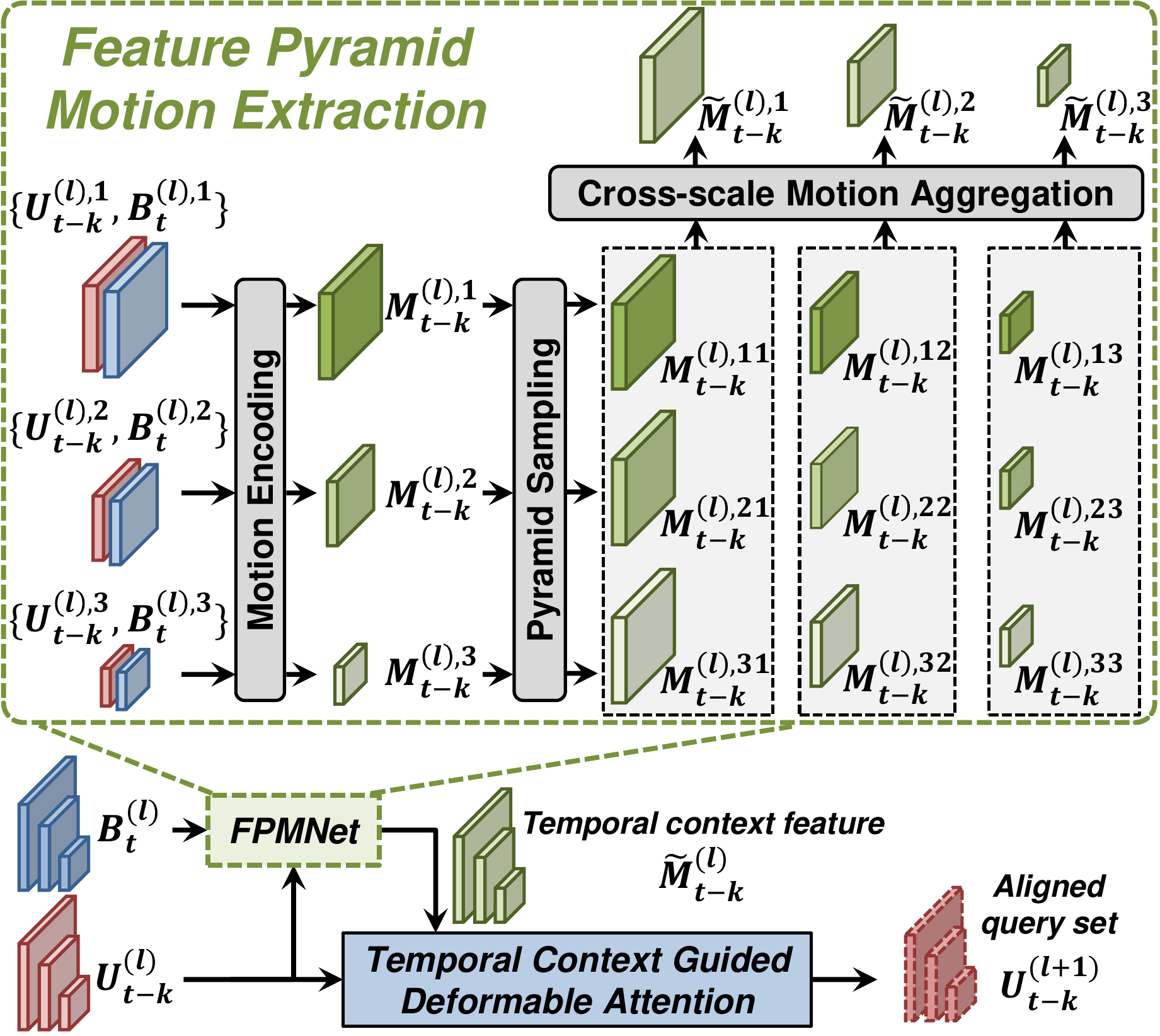}}
    	\caption {\textbf{Structure of IDANet:} IDANet aligns S-QS using deformable attention. The temporal context information extracted from FPMNet is used to predict the sampling offsets and attention weights required for deformable attention.}
	\label{Figure_Deformable_Self_Alignment}
\end{figure}

\subsection{Inter-frame deformable alignment network (IDANet)}
 Fig. \ref{Figure_Deformable_Self_Alignment} depicts the structure of IDANet. First, FPMNet extracts temporal motion features from the adjacent multi-scale feature maps. The proposed deformable attention performs an inter-frame feature alignment based on these temporal features.

\textbf{Feature pyramid motion extraction:}
The FPMNet operates on the dual query sets $\textbf{\textit{B}}^{(l)}_t$ and $\textbf{\textit{U}}^{(l)}_{t-k}$ to produce the temporal context features $\tilde{\textbf{\textit{M}}}^{(l)}_{t-k}$. Recall that $\textbf{\textit{B}}^{(l)}_t$ and $\textbf{\textit{U}}^{(l)}_{t-k}$ represent the target frame features and the $(t-k)$th support frame features updated  up to the $l$th attention layer. 
First, FPMNet applies the motion encoding to $\textbf{\textit{B}}^{(l)}_t$ and $\textbf{\textit{U}}^{(l)}_{t-k}$ for each scale, i.e., 
\begin{align}
    \textbf{\textit{M}}^{(l),s}_{t-k} = \mathrm{conv}_{3\times3} (\textbf{\textit{B}}^{(l),s}_{t}-\textbf{\textit{U}}^{(l),s}_{t-k}),
\end{align}
where  ${\mathrm{ conv_{3\times3}}}$ denotes the convolutional layer with $3\times3$ kernels and $\textbf{\textit{M}}^{(l),s}_{t-k}$, $\textbf{\textit{B}}^{(l),s}_{t}$, and $\textbf{\textit{U}}^{(l),s}_{t-k}$ are the scale-$s$ features from  $\textbf{\textit{M}}^{(l)}_{t-k}$, $\textbf{\textit{B}}^{(l)}_{t}$, and $\textbf{\textit{U}}^{(l)}_{t-k}$, respectively. We enhance the temporal context features further by capturing the multi-scale contexts across all scales. As shown in Fig. \ref{Figure_Deformable_Self_Alignment}, we use pyramid sampling \cite{TANet} to transform the features $\textbf{\textit{M}}^{(l),s}_{t-k}$ of the scale $s$ into the features $\textbf{\textit{M}}^{(l),ss'}_{t-k}$ of the scale $s'$. Then, the final temporal context features $\tilde{\textbf{\textit{M}}}^{(l),s'}_{t-k}$ are obtained by aggregating the set of the $s'$-scale features  $\{\textbf{\textit{M}}^{(l),ss'}_{t-k}\}_{s'=1}^{S}$, i.e.,  
\begin{align}
    \tilde{\textbf{\textit{M}}}^{(l),s'}_{t-k} = \mathrm{conv}_{1\times1} ([\textbf{\textit{M}}^{(l),1s'}_{t-k},\textbf{\textit{M}}^{(l),2s'}_{t-k},...,\textbf{\textit{M}}^{(l),Ss'}_{t-k}]),
\end{align}
where $[\cdot,\cdot,...,\cdot]$ denotes the channel-wise concatenation.
The detailed process of extracting the temporal context features is described in Fig. 3.
Finally, FPMNet produces the multi-scale temporal context features  $\tilde{\textbf{\textit{M}}}^{(l)}_{t-k} = \{\tilde{\textbf{\textit{M}}}^{(l),s'}_{t-k}\}_{s'=1}^{S}$. Note that these feature maps are used to predict both the mask offsets and weights of the deformable attention.

\textbf{Temporal context guided deformable attention:}
For each attention layer, IDANet applies the deformable attention to S-QS, $\textbf{\textit{U}}^{(l)}_{t-k}$ and produces the refined S-QS,  $\textbf{\textit{U}}^{(l+1)}_{t-k}$. We consider the deformable attention, which supports $H$ multi-head attentions. Let $p_s$ be a two-dimensional reference position on the feature map of the scale $s$. Note that the reference position $p_s$ is swept over the entire feature map to compute the attention values.
The BEV embedding feature $\textbf{\textit{V}}^{sh}_{t-k}$ for the $h$th head and the $s$th scale is obtained as
\begin{align}
    \textbf{\textit{V}}^{sh}_{t-k}(p_s) = W_{h}\, \textbf{\textit{U}}^{(l),s}_{t-k}(p_s),
\end{align}
where  $W_{h}\in\mathbb{R}^{D' \times D}$ is the projection matrix and $D'$ is $(D/H)$.

Next, the mask offsets $\Delta_{t-k}^{sh}(p_s)$ and the attention weights $A_{t-k}^{sh}(p_s)$ at the reference point $p_s$  are given by 
\begin{align}
    \Delta_{t-k}^{sh}(p_s) &= W'_{h}\, \tilde{\textbf{\textit{M}}}^{(l),s}_{t-k}(p_s) \\
    A_{t-k}^{sh}(p_s) &= \mathrm{softmax}(W''_{h}\, \tilde{\textbf{\textit{M}}}^{(l),s}_{t-k}(p_s)),
\end{align}
where $W'_{h}\in\mathbb{R}^{(2  S  J) \times D}$ and $W''_h \in\mathbb{R}^{(S J) \times D}$ are the linear projection matrices, $J$ is the size of the mask, and $\mathrm{softmax}(\cdot)$ is the softmax function.
Note that $\Delta_{t-k}^{sh}(p_s)$ is the set $\{\Delta_{t-k}^{shij}(p_s) \vert 1\leq i\leq S, 1\leq j\leq J \}$, where 
$\Delta_{t-k}^{shij}(p_s)$ is the $(x,y)$ mask offset of the $j$th sampling point at the $i$th scale.
$A_{t-k}^{sh}(p_s)$ is defined similarly as $\{A_{t-k}^{shij}(p_s) \vert 1\leq i\leq S, 1\leq j\leq J \}$, except that each element has a scalar weight.

Given the mask offsets and weights,  the attention value $z^s_{t-k}(p_s)$ can be obtained from
\begin{align}
    &z^s_{t-k}(p_s)=\sum_{h=1}^{H}W'''_{h} \cdot \nonumber \\
    & \quad \left(\sum_{i=1}^{S}\sum_{j=1}^{J}A_{t-k}^{shij}(p_s)\cdot\textbf{\textit{V}}^{ih}_{t-k}\left(\psi_i{(p_s)}+\Delta_{t-k}^{shij}(p_s)\right)\right),
\end{align}
where $W'''_h\in\mathbb{R}^{D \times D'}$ is the projection matrix
and $\psi_i(\cdot)$ is a re-scaling function that adjusts the reference position $p_s$ for the $i$th scale.

Finally, the refined S-QS $\textbf{\textit{U}}^{(l+1),s}_{t-k}$ is obtained as
\begin{align}
    \textbf{\textit{U}}^{(l+1),s}_{t-k}(p_s) = \mathrm{FFN}(\mathrm{LN}(\mathrm{dropout}(z^s_{t-k}(p_s)) + \textbf{\textit{U}}^{(l),s}_{t-k}(p_s))),
\end{align}
where $\mathrm{FFN}(\cdot)$ is the feed-forward network \cite{def-detr}, $\mathrm{dropout}(\cdot)$ is the dropout \cite{dropout}, and $\mathrm{LN}(\cdot)$ is the layer normalization \cite{layer_norm}.

\subsection{Inter-frame gated aggregation network (IGANet)}
 IGANet aggregates S-QS $\{\textbf{\textit{U}}^{(l+1)}_{t-k}\}_{k=1}^{N-1}$ with T-QS $\textbf{\textit{B}}^{(l)}_t$ to produce the refined T-QS $\textbf{\textit{B}}^{(l+1)}_{t}$.
 IGANet applies the gated attention network \cite{GAM} to combine $\textbf{\textit{U}}^{(l+1)}_{t-k}$ and $\textbf{\textit{B}}^{(l)}_t$  for each scale
\begin{align}
    \tilde{\textbf{\textit{B}}}^{(l),s}_{t-k} = \textbf{\textit{G}}^{(l),s}_{t-k}\otimes \textbf{\textit{B}}^{(l),s}_t + (1-\textbf{\textit{G}}^{(l),s}_{t-k})\otimes \textbf{\textit{U}}^{(l+1),s}_{t-k},
\end{align}
where the operation $\otimes$ denotes element-wise multiplication and $\textbf{\textit{G}}^{(l),s}_{t-k}$ denotes the gating matrix that weights the contribution from S-QS and T-QS. The gating weights $\textbf{\textit{G}}^{(l),s}_{t-k}$ are obtained by applying the convolutional layers to the concatenation of S-QS and T-QS,
\begin{align}
    \textbf{\textit{G}}^{(l),s}_{t-k} = \sigma(\mathrm{conv_{3\times3}}([\textbf{\textit{B}}^{(l),s}_{t},\textbf{\textit{U}}^{(l+1),s}_{t-k}]) ,
\end{align}
where $\sigma(\cdot)$ is the sigmoid function.
Note that the gating matrix $\textbf{\textit{G}}^{(l),s}_{t-k}$ has a channel dimension of one and each element has a value between 0 and 1.

Finally, we aggregate the outputs $\tilde{\textbf{\textit{B}}}^{(l),s}_{t-1},..., \tilde{\textbf{\textit{B}}}^{(l),s}_{t-N+1}$ of the gated attention network over frames and obtain the refined T-QS as
\begin{align}
    \textbf{\textit{B}}^{(l+1),s}_{t} = \mathrm{conv_{3\times3}}([\tilde{\textbf{\textit{B}}}^{(l),s}_{t-1},..., \tilde{\textbf{\textit{B}}}^{(l),s}_{t-N+1}]).
\end{align}
Putting $\textbf{\textit{B}}^{(l+1),s}_{t}$ together for all scales, IGANet produces the refined T-QS, $\textbf{\textit{B}}^{(l+1)}_{t}$.

\newcolumntype{C}{>{\centering\arraybackslash}p{2.3em}}

\renewcommand{\arraystretch}{1.0}

\begin{table*}[t]\caption{Quantitative comparison with state of the art methods on nuScenes test set.
C.V and T.C presents the construction vehicle and the traffic cone, respectively. Motor. and Ped. are short for the motorcycle and the pedestrian, respectively.
L indicates the LiDAR. The best performance is bolded.}
\begin{center}

\begin{adjustbox}{width=0.95\linewidth}
{\normalsize
\begin{tabular}{c | c | C  C |  C  C  C  C  C  C  C  C  C  C }
\Xhline{4\arrayrulewidth}


Method & Input & NDS & mAP & Car & Truck & Bus & Trailer & C.V & Ped. & Motor. & Bicycle & T.C & Barrier \\ \hline\hline


PointPillars \cite{pointpillars}& \multirow{12}{*}{\begin{tabular}[c]{@{}c@{}}Single\\frame\end{tabular}} & 45.3 & 30.5 & 68.4 & 23.0 & 28.2 & 23.4 & 4.1 & 59.7 & 27.4 & 1.1 & 30.8 & 38.9\\
3DSSD \cite{3dssd} &  & 56.4 & 42.6 & 81.2 & 47.2 & 61.4 & 30.5 & 12.6 & 7.2 & 36.0 & 8.6 & 31.1 & 47.9\\
SA-Det3D \cite{sa-det3d} &  & 59.2 & 47.0 & 81.2 & 43.8 & 57.2 & 47.8 & 11.3 & 73.3 & 32.1 & 7.9 & 60.6 & 55.3\\
SSN V2 \cite{ssnV2} &  & 61.6 & 50.6 & 82.4 & 41.8 & 46.1 & 48.0 & 17.5 & 75.6 & 48.9 & 24.6 & 60.1 & 61.2\\
CBGS \cite{cbgs} &  & 63.3 & 52.8 & 81.1 & 48.5 & 54.9 & 42.9 & 10.5 & 80.1 & 51.5 & 22.3 & 70.9 & 65.7\\
CVCNet \cite{cvcnet} &  & 64.2 & 55.8 & 82.7 & 46.1 & 45.8 & 46.7 & 20.7 & 81.0 & 61.3 & 34.3 & 69.7 & 69.9\\
CyliNet \cite{cylinet} &  & 66.1 & 58.5 & 85.0 & 50.2 & 56.9 & 52.6 & 19.1 & 84.3 & 58.6 & 29.8 & 79.1 & 69.0\\
HotSpotNet \cite{hotspotnet} &  & 66.6 & 59.3 & 83.1 & 50.9 & 56.4 & 53.3 & 23.0 & 81.3 & 63.5 & 36.6 & 73.0 & 71.6\\
CenterPoint \cite{centerpoint} &  & 67.3 & 60.3 & 85.2 & 53.5 & 63.6 & 56.0 & 20.0 & 84.6 & 59.5 & 30.7 & 78.4 & 71.1\\
AFDet V2 \cite{afdetv2} &  & 68.5 & 62.4 & \bf 86.3 & 54.2 & 62.5 & 58.9 & 26.7 & \bf85.8 & 63.8 & 34.3 & 80.1 & 71.0 \\
S2M2-SSD \cite{s2m2ssd} &  & 69.3 & 62.9 & \bf 86.3 & 56.0 & 65.4 & 59.8 & 26.2 & 84.5 & 61.6 & 36.4 & 77.7 & \bf 75.1\\
UVTR-L \cite{UVTR} &  & 69.7 & 63.9 & \bf 86.3 & 52.2 & 62.8 & 59.7 & \bf 33.7 & 84.5 & 68.8 & 41.1 & 74.7 & 74.9\\


\hline
3DVID  \cite{3dvid} & \multirow{5}{*}{\begin{tabular}[c]{@{}c@{}}Multi\\frame\end{tabular}}    & 53.1 & 45.4 & 79.7 & 33.6 & 47.1 & 43.1 & 18.1 & 76.5 & 40.7 & 7.9 & 58.8 & 48.8\\
TCTR \cite{tctr}     &   &  \textbf{-}    & 50.5 & 83.2 & 51.5 & 63.7 & 33.0 & 15.6 & 74.9 & 54.0 & 22.6 & 52.5 & 53.8\\
STGNN \cite{stgnn}   &   & 66.7 & 59.0 & 86.2 & \bf 57.2 & \bf 67.2 & 35.5 & 14.6 & 85.5 & 58.1 & 37.0 & 71.3 & 66.2\\
Our D-Align-PP           &   & 63.1 & 53 & 82.2 & 44.0 & 53.1 & 53.5 & 19.8 & 75.8 & 51.3 & 17.6 & 67.4 & 65.5\\
Our D-Align-CP           &   & \bf 70.2 & \bf 64.0 & 85.0 & 50.5 &  67.0 & \bf 60.2 & 27.5 & 84.7 & \bf 73.6 & \bf 43.3 & \bf 81.8 & 66.0\\

\Xhline{4\arrayrulewidth}

\end{tabular}
}
\end{adjustbox}
\end{center}

\label{table:sota_new}
\end{table*}
\renewcommand{\arraystretch}{1}

\section{EXPERIMENTS}
\subsection{nuScenes Dataset}
The nuScenes dataset \cite{nuscenes} is a challenging large-scale autonomous driving dataset comprising 1000 scenes, each with a duration of about 20 seconds. The dataset provides 32-channel LiDAR scan samples acquired at 20Hz with a full 360-degree view. Moreover, 3D bounding boxes from 10 categories are annotated at 2Hz, i.e.,  annotated key samples are supplied in every 10 samples. We evaluated our method in terms of mean average precision (mAP) and nuScenes detection score (NDS), which are the official 3D object detection benchmark metrics of nuScenes.
The mAP is computed with the 2D center distance at the bird's-eye-view between the ground-truth data and the predictions. The NDS is a weighted sum of mAP and several true positive metrics, i.e., the average translation, scale, orientation, velocity, and attribute errors.

\subsection{Implementation Details}
{\bf Point Cloud Sequence Processing.} 
Following the nuScenes dataset \cite{nuscenes}, a single frame comprised $10$ consecutive LiDAR point cloud samples $(T=0.5)$ with ego-vehicle motion compensation applied. Specifically, each point in the single frame is represented with 5 dimensional vector as $(x,y,z,r,\Delta t)$, where $r$ is intensity of the point and $\Delta t$ indicates the time lag to the key sample, i.e., a scalar time stamp in the range of [0, 0.5). Unless otherwise stated, $N=3$ frames were used, in which the input sequence consists of a single target frame and two previous support frames.

\textbf{Architecture.}
We constructed the proposed D-Align based on two types of 3D object detector baselines, including anchor-based and anchor-free detectors.
First, we chose PointPillars \cite{pointpillars} as an anchor-based detector baseline. The point cloud range was set to $[-51.2,51.2]\times[-51.2,51.2]\times[-5.0,3.0]$ meters on the $(x,y,z)$ axes, and the grid size was $0.2 \times 0.2 \times 8.0m$.
The PointPillars baseline generated the BEV feature maps of $S=3$ scales, the sizes of which were $(D, H/2, W/2)$, $(D, H/4, W/4)$, and $(D, H/8, W/8)$. $(H, W)$ denotes the size of the voxel grid structure in the $x$-$y$ axis. Next, we chose CenterPoint \cite{centerpoint} with a VoxelNet \cite{voxelnet} backbone as an anchor-free detector baseline. The point cloud range was $[-54.0,54.0]\times[-54.0,54.0]\times[-5.0,3.0]$ meters on the $(x,y,z)$ axes, and the grid size was $0.075 \times 0.075 \times 0.2m$. It generated the BEV feature maps of $S=2$ scales and their sizes were set to $(D, H/8, W/8)$ and $(D, H/16, W/16)$. In DUCANet, the number of attention layers was set to $L=3$. In the deformable attention of IDANet, the number of attention heads $H$ and sampling locations $J$ are set to 8 and 4, respectively. Dropout was applied with a ratio of 0.1, as in \cite{Transformer}.

{\bf Training.}
We adopted the loss functions used in PointPillars \cite{pointpillars} and CenterPoint \cite{centerpoint} to train D-Aligns and used the Adam optimizer. D-Align was trained in two steps: 1) pre-training and 2) fine-tuning. In pre-training step, we trained a single frame detector including only the BEV feature extraction backbone and the 3D detection head. We used the one-cycle {\it learning rate} (LR) scheduling for 20 epochs with a maximum LR of 0.001. Then, we fine-tuned the entire network with the pre-trained weights initialized. In fine-tuning, we used the same LR scheduler for 40 epochs with a maximum LR of 0.0002. The batch size was set to 16 for pre-training and 8 for fine-tuning. We applied data augmentation methods in both steps, including random flipping, rotation, scaling, and ground truth box sampling \cite{second}.

\subsection{Detection Performance on nuScenes $test$ set}
We evaluated the performance of our proposed D-Align in comparison with other LiDAR-based 3D object detectors. We trained our model on the entire nuScenes training set and evaluated its 3D object detection performance on the provided testing set. For convenience, we refer to D-Align with a PointPillars \cite{pointpillars} baseline as {\it D-Align-PP} and D-Align with a CenterPoint \cite{centerpoint} baseline as  {\it D-Align-CP}. In Table \ref{table:sota_new}, we report the test detection accuracy for both D-Align-PP and D-Align-CP. We did not use any ensemble strategies. Other 3D object detectors include both single-frame methods and multi-frame methods. D-Align-CP achieved state-of-the-art (SOTA) performance, outperforming the latest 3D object detectors in terms of both mAP and NDS. D-Align exhibited significant improvements in performance over the single-frame based baselines. The D-Align-PP and D-Align-CP achieved 22.5\% gain over PointPillar baseline and 3.7\% gain in mAP  over CenterPoint baseline, which shows that the proposed method effectively utilized the spatio-temporal information in the point cloud sequence. D-Align performed particularly well in motorcycle, bicycle, and traffic cone object classes, which tend to be challenging to detect due to the lack of LiDAR points. D-Align compensates for the performance degradation caused by a lack of points by aggregating information from multi-frame features.

\renewcommand{\arraystretch}{1.2}

\begin{table}[t]
\caption{Ablation studies conducted on nuScnenes validation set}
\label{table:ablation}
\begin{center}
\begin{adjustbox}{width=0.45\textwidth}
{\normalsize
\begin{tabular}{c|c|c|c|c|c}
\Xhline{4\arrayrulewidth}
\multicolumn{2}{c|}{\multirow{2}{*}{}}& \multicolumn{4}{c}{Methods} \\ \cline{3-6} 
\multicolumn{2}{c|}{} & (a) & (b) & (c) & (d) \\ \hline 

\multicolumn{2}{c|}{IGANet} & & \checkmark & \checkmark & \checkmark \\ \cline{1-2} 
\multirow{2}{*}{\begin{tabular}[c]{@{}c@{}}IDANet\end{tabular}} & w/o FPMNet & &  & \checkmark & \\ \cline{2-2}
& w/ FPMNet & & & & \checkmark \\ \hline

\multicolumn{2}{c|}{mAP [\%]} & 44.40 & 46.17 & 48.19 & 49.66 \\ \hline
\multicolumn{2}{c|}{NDS [\%]} & 58.15 & 59.42 & 60.60 & 61.33 \\

\Xhline{4\arrayrulewidth}
\end{tabular}
}

\end{adjustbox}
\end{center}
\end{table}
\renewcommand{\arraystretch}{2}
\renewcommand{\arraystretch}{1.2}

\begin{table}[t]
\caption{Performance vs. the number of layers}
\label{table:Num_ATTN_Layers}
\begin{center}
\begin{adjustbox}{width=0.45\textwidth}
{\tiny

\begin{tabular}{c|c|c|c}
\Xhline{2\arrayrulewidth} 
 & \multicolumn{3}{c}{D-Align}\\ \hline 
\# of attention layers & 1 & 2 & 3 \\ \hline 

mAP [\%] & 46.44 & 48.58 & 49.66 \\ \hline
NDS [\%] & 59.55 & 60.72 & 61.33 \\

\Xhline{2\arrayrulewidth}

\end{tabular}
}
\end{adjustbox}
\end{center}
\end{table}

\subsection{Ablation Studies on nuScenes $validation$ set}
We conducted ablation studies on the nuScenes validation set to confirm the effectiveness of each component of D-Align.
PointPillars \cite{pointpillars} was used as a baseline detector and was trained with the whole nuScenes training set.
We fine-tuned the entire D-Align network only with 1/7 training set to reduce the training effort.

\textbf{Contribution of D-Align components.}
Table \ref{table:ablation} provides the contributions of IGANet, IDANet, and FPMNet to overall detection performance.
Method (a) is the single-frame-based PointPillars baseline. Method (b) adds only IGANet to the baseline detector. The multi-frame features are not aligned and are directly aggregated by IGANet. This improves the performance of Method (a) by 1.77\% and 1.27\% in mAP and NDS, respectively. The performance improvements that can be achieved through feature aggregation are limited due to non-negligible discrepancy between multi-frame features. Method (c) adds inter-frame feature alignment by IDANet to Method (b) without using FPMNet. Instead of using the temporal context features, T-QS is used to determine the mask offsets and weights.
By aligning the multi-frame features through IDANet, Method (c) improves performance by 2.02\% in mAP and 1.18\% in NDS compared to Method (b).
Finally, Method (d) is the proposed D-Align architecture with FPMNet. Method (d) yields additional improvements of 1.47\% and 0.73\% in mAP and NDS over Method (c), respectively. This shows that the temporal context produced by FPMNet helps achieve better attention to multi-frame features.

\begin{figure}[t]
    \centering
    \begin{subfigure}[mAP Performance]{\includegraphics[width=0.47\columnwidth]{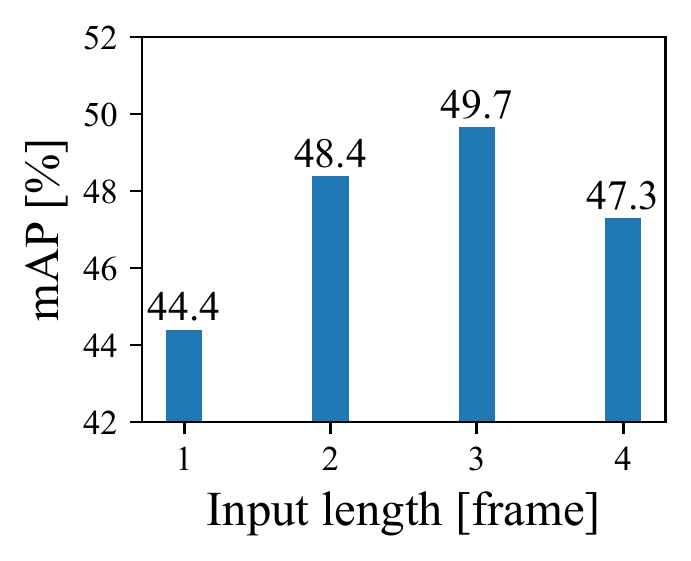}}
    \end{subfigure}
    \hspace{0.3mm}
    \begin{subfigure}[NDS Performance]{\includegraphics[width=0.47\columnwidth]{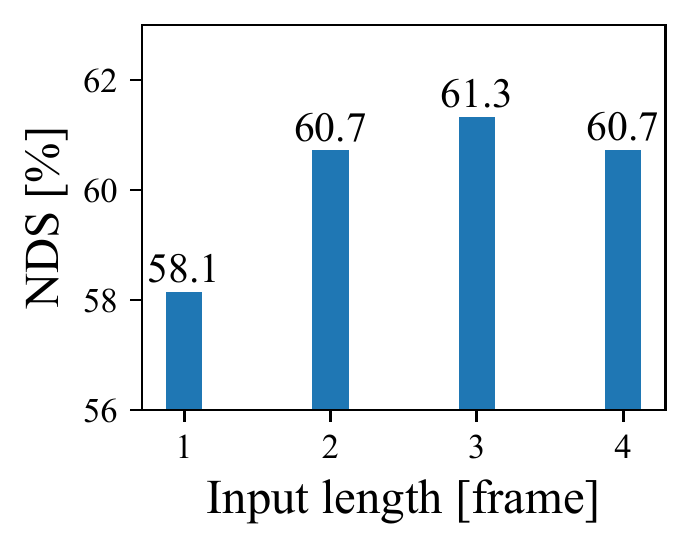}}
    \end{subfigure}
    \caption {\textbf{Performance vs. input frame length}}
    \label{Ablation_seq_length}
\end{figure}

\textbf{Performance versus the number of attention layers.}
Table \ref{table:Num_ATTN_Layers} presents the performance of D-Align as a function of the number of attention layers, $L$. Due to the memory limitation, we increased $L$ only up to  three.
The detection accuracy of D-Align improves with $L$. This confirms that the dual query attention progressively improves the BEV representation through successive layers of attention.  

\textbf{Performance versus length of input frames.}
Fig. \ref{Ablation_seq_length} provides performance change with respect to the length of the input frames.
The accuracy of D-Align improves up to an input frame length of 3, but this trend stops at a length of 4. Considering that 4 frames in the nuScenes dataset are equivalent to 2 seconds, it seems difficult for D-Align to cope with dynamic changes in features that occur in durations over 2 seconds.

\section{CONCLUSIONS}
In this paper, we have proposed a novel dual-query based co-attention method called D-Align for 3D object detection, which can leverage spatio-temporal information in point cloud sequences.
The proposed dual-query framework uses two query sets, T-QS and S-QS, which serve to refine the features of both the target and support frames in an iterative fashion.
We designed IDANet to align S-QS to T-QS based on the temporal context features generated by FPMNet.
We also introduced IGANet, which can aggregate dual query sets to refine T-QS. IDANet and IGANet transform the input query sets and produce the updated queries in each attention layer. Finally, the proposed D-Align produces the enhanced target frame BEV feature maps for 3D object detection. We have also evaluated the performance of D-Align on public nuScenes dataset. 
Our experimental results have confirmed that the proposed D-Align exhibited substantial performance improvements compared to the baseline 3D object detectors and achieved state-of-the-art performance among the latest LiDAR-based 3D object detectors.

\bibliographystyle{plain}
\bibliography{Reference}

\end{document}